\documentclass[11pt,a4paper]{article}

\usepackage{naaclhlt2019}
\usepackage{times}
\usepackage{latexsym}

\usepackage{epstopdf}
\usepackage{multirow}
\usepackage{graphicx}

\newcommand{\shadowed}[1]{\textcolor{gray}{#1}}
\renewcommand{\vec}[1]{\mathbf{#1}}
\usepackage{amsmath}

\usepackage{cleveref} %best refs

\aclfinalcopy % Uncomment this line for the final submission
\title{Better, Faster, Stronger Sequence Tagging Constituent Parsers}

\author{David Vilares \\
  Universidade da Coru\~na, CITIC \\
  Departamento de Computaci\'on \\
  A Coru\~na, Spain  \\
  {\tt david.vilares@udc.es} \\\And
  Mostafa Abdou, Anders S\o gaard \\
  University of Copenhagen \\
  Department of Computer Science \\
  Copenhagen, Denmark \\
  {\tt \{abdou,soegaard\}@di.ku.dk} \\}

\date{}

\begin{document}
\maketitle

\begin{abstract}

Sequence tagging models for constituent parsing are faster, but less accurate than other types of parsers. In this work, we address the following weaknesses of such constituent parsers: (a) high error rates around closing brackets of long constituents, (b) large label sets, leading to sparsity, and (c) error propagation arising from greedy decoding. To effectively close brackets, we train a model that learns to switch between tagging schemes. To reduce sparsity, we decompose the label set and use multi-task learning to jointly learn to predict sublabels. Finally, we mitigate issues from greedy decoding through auxiliary losses and sentence-level fine-tuning with policy gradient. Combining these techniques, we clearly surpass the performance of sequence tagging constituent parsers on the English and Chinese Penn Treebanks, and  reduce their parsing time even further. On the \textsc{spmrl} datasets, we observe even greater improvements across the board, including a new state of the art on Basque, Hebrew, Polish and Swedish.\footnote{After this paper was submitted, \newcite{Kitaev2018BERT} have improved our results using their previous self-attentive constituent parser \cite{KitaevConstituencyACL2018} and \textsc{bert} representations \cite{devlin2018bert} as input to their system. We will acknowledge these results in the Experiments section.}

\end{abstract}

\noindent\fbox{%
    \parbox{\columnwidth}{%
This is a revised version of the paper originally published in NAACL 2019, with a corrigendum at the end describing the changes. The previous version contained a bug where the script \textsc{evalb} for comparison against the state-of-the-art was not considering the .prm parameter files. 
    }%
}
\vspace{0.2cm}

\section{Introduction}

Constituent parsing is a core task in natural language processing (\textsc{nlp}), with a wide set of applications. Most competitive parsers are slow, however, to the extent that it is prohibitive of downstream applications in large-scale environments \cite{kummerfeldCorral}. Previous efforts to obtain speed-ups have focused on creating more efficient versions of traditional shift-reduce \cite{sagae2006best,zhang2009transition} or chart-based parsers \cite{collins1997three,charniak2000maximum}.
\newcite{zhu2013fast}, for example, presented a fast shift-reduce parser with transitions learned by a \textsc{svm} classifier. Similarly, \newcite{hall2014less} introduced a fast \textsc{gpu} implementation for \newcite{petrov2007improved}, and \newcite{ShenDistance2018} significantly improved the speed of the \newcite{stern2017minimal} greedy top-down algorithm, by learning to predict a list of syntactic distances that determine the order in which the sentence should be split.

In an alternative line of work, some authors have proposed new parsing paradigms that aim to both reduce the complexity of existing parsers and improve their speed. \newcite{vinyals2015grammar} proposed a machine translation-inspired sequence-to-sequence approach to constituent parsing, where the input is the raw sentence, and the `translation' is a parenthesized version of its tree. \newcite{GomVilEMNLP2018} reduced constituent parsing to sequence tagging, where only $n$ tagging actions need to be made, and obtained one of the fastest parsers to date. However, the performance is well below the state of the art \cite{DyerRecurrent2016,stern2017minimal,KitaevConstituencyACL2018}.

\paragraph{Contribution} We first explore different factors that prevent sequence tagging constituent parsers from obtaining better results. These include: high error rates when long constituents need to be closed, label sparsity, and error propagation arising from greedy inference. We then present the technical contributions of the work. To effectively close brackets of long constituents, we combine the relative-scale tagging scheme used by \newcite{GomVilEMNLP2018} with a secondary top-down absolute-scale scheme. This makes it possible to train a model that learns how to switch between two encodings, depending on which one is more suitable at each time step. To reduce label sparsity, we recast the constituent-parsing-as-sequence-tagging problem as multi-task learning ({\sc mtl}) \cite{caruana1997multitask}, to decompose a large label space and also obtain speed ups. Finally, we mitigate error propagation using two strategies that come at no cost to inference efficiency: auxiliary tasks and policy gradient fine-tuning.

\section{Preliminaries}

We briefly introduce preliminaries that we will build upon in the rest of this paper: encoding functions for constituent trees, sequence tagging, multi-task learning, and reinforcement learning.

\paragraph{Notation} We use $w$=$[w_0,w_1,...,w_n]$ to refer to a raw input sentence and bold style lower-cased and math style upper-cased characters to refer to vectors and matrices, respectively (e.g. $\vec{x}$ and $\vec{W}$).

\subsection{Constituent Parsing as Sequence Tagging}\label{section-linearization-of-trees}

\newcite{GomVilEMNLP2018} define a linearization function of the form $\Phi_{|w|}: T_{|w|} \rightarrow L^{(|w|-1)}$ to map a phrase structure tree with $|w|$ words to a sequence of labels of length $|w|-1$.\footnote{They (1) generate a dummy label for the last word and (2) pad sentences with a beginning- and end-of-sentence tokens.} For each word $w_t$, the function generates a label $l_t \in L$ of the form $l_t$=$(n_t,c_t,u_t)$, where:

\begin{itemize}
\item $n_t$ encodes the number of ancestors in common between between $w_t$ and $w_{t+1}$. To reduce the number of possible values, $n_t$ is encoded as the relative variation in the number of common ancestors with respect to $n_{t-1}$.
\item $c_t$ encodes the lowest common ancestor between $w_t$ and $w_{t+1}$.
\item $u_t$ contains the unary branch for $w_t$, if any.
\end{itemize}

\begin{figure}[hbtp]
\centering
\includegraphics[width=1\columnwidth]{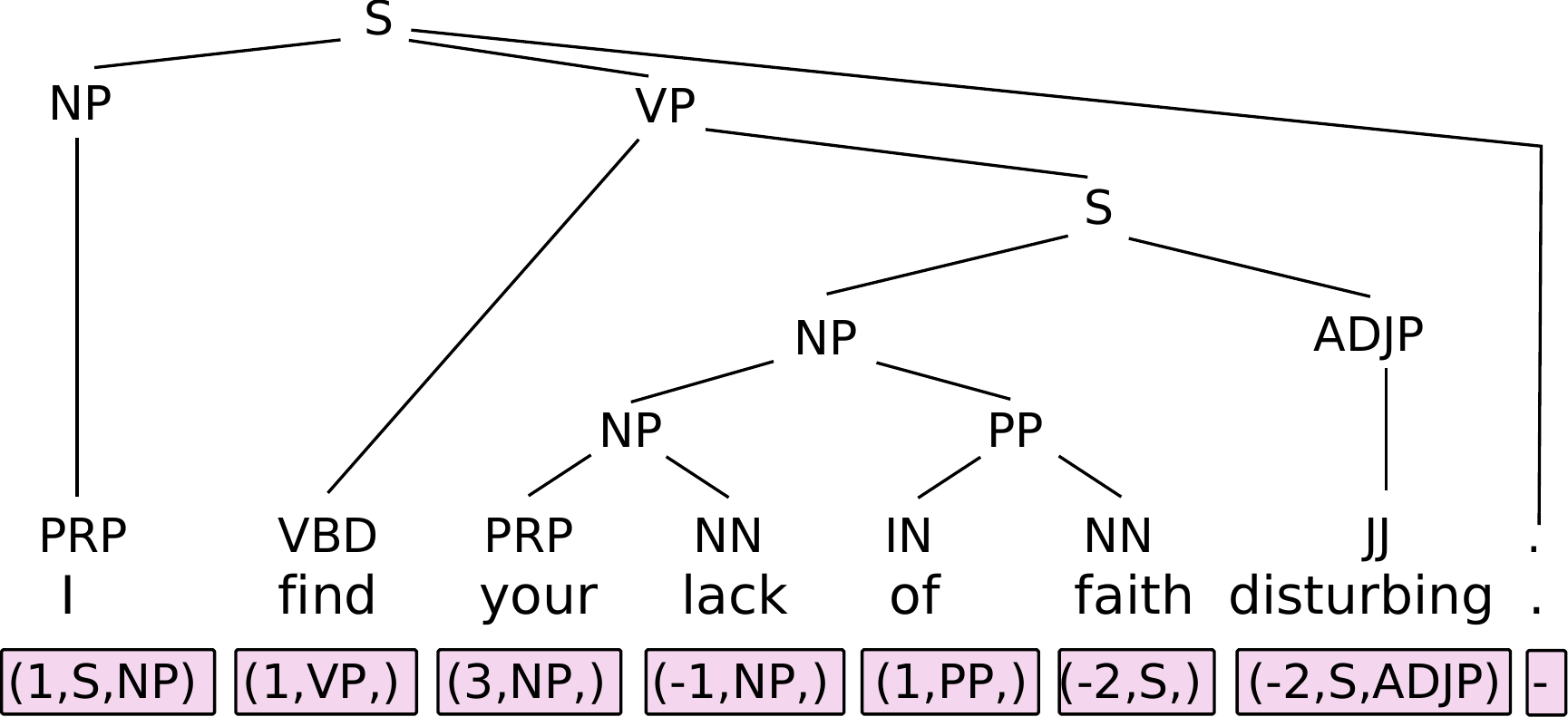}
\caption{\label{f-seq-lab-example} A constituent tree linearized as by \newcite{GomVilEMNLP2018}.}
\end{figure}

Figure \ref{f-seq-lab-example} explains the encoding with an example.

\subsection{Sequence Tagging}\label{section-sequence-tagging}

Sequence tagging is a structured prediction task that generates an output label for every input token. Long short-term memory networks (\textsc{lstm}) \cite{hochreiter1997long} are a popular architecture for such tasks, often giving state-of-the-art performance  \cite{Reimers:2017:EMNLP,yang2017ncrf}.

\begin{figure}[hbtp]
\centering
\includegraphics[width=1\columnwidth]{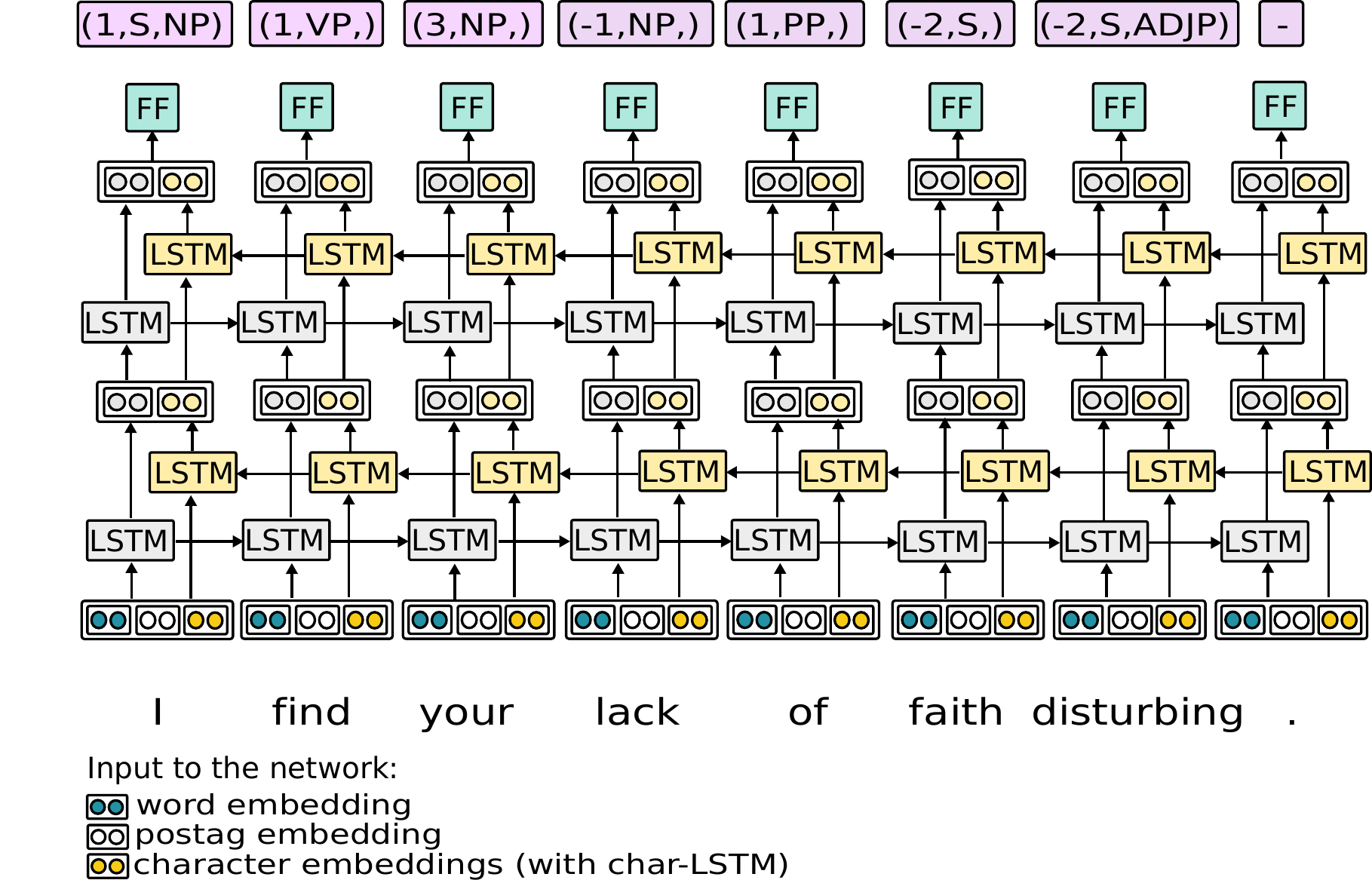}
\caption{\label{f-baseline-architecture} The baseline architecture used in this work. The input to the network is a concatenation of word embeddings, PoS-tag embeddings and a second word embedding learned by a character-based \textsc{lstm} layer.}
\end{figure}

\paragraph{Tagging with \textsc{lstm}s} In \textsc{lstm}s, the prediction for the $i$th element is conditioned on the output of the previous steps. Let \textsc{lstm}$_\theta$($\vec{x}_{1:n}$) be a parametrized function of the network, where the input is a sequence of vectors $\vec{x}_{1:n}$, its output is a sequence of hidden vectors $\vec{h}_{1:n}$. To obtain better contextualized hidden vectors, it is possible to instead use bidirectional \textsc{lstms} \cite{schuster1997bidirectional}. First, a \textsc{lstm}$_\theta^l$ processes the tokens from left-to-right and then an independent \textsc{lstm}$_\theta^r$ processes them from right-to-left. The $i$th final hidden vector is represented as the concatenation of both outputs, i.e.
\textsc{bilstm}$_\theta(\vec{x},i)$ = $\textsc{lstm$_\theta^l$}(\vec{x}_{[1:i]}) \circ \textsc{lstm$_\theta^r$}(\vec{x}_{[|\vec{x}|:i]})$. 
\textsc{bilstm}s can be stacked in order to obtain richer representations. To decode the final hidden vectors into discrete labels, a standard approach is to use a feed-forward network together with a softmax transformation, i.e. $P(y|\vec{h}_i)$ = $softmax(W \cdot \vec{h}_i + \vec{b})$. 
We will use the \textsc{bilstm}-based model by \newcite{yang2017ncrf}, for direct comparison against \newcite{GomVilEMNLP2018}, who use the same model.
As input, we will use word embeddings, PoS-tag embeddings and a second word embedding learned by a character-based \textsc{lstm} layer. The model is optimized minimizing the categorical cross-entropy loss, i.e. $\mathcal{L}$ = $-\sum{log(P(y|\vec{h}_i))}$. The architecture is shown in Figure \ref{f-baseline-architecture}.

\subsection{Multi-task Learning}\label{section-multitask}

Multi-task learning is used to solve multiple tasks 
using a single model architecture, with task-specific classifier functions from the outer-most representations \cite{caruana1997multitask,collobert2008unified}. The benefits are intuitive: sharing a common representation for different tasks acts as a generalization mechanism and allows to address them in a parallel fashion. The \emph{hard-sharing} strategy is the most basic \textsc{mtl} architecture, where the internal representation is fully shared across all tasks. The approach has proven robust for a number of \textsc{nlp} tasks \cite{Bingel:ea:17} and comes with certain guarantees if a common, optimal representation exists \cite{Baxter:00}.
\newcite{dong2015multi} use it for their multilingual machine translation system,  where the encoder is a shared gated recurrent neural network \cite{cho2014properties} and the decoder is language-specific. \newcite{plank2016multilingual} also use a hard-sharing setup to improve the performance of \textsc{bilstm}-based PoS taggers. To do so, they rely on \emph{auxiliary tasks}, i.e, tasks that are not of interest themselves, but that are co-learned in a \textsc{mtl} setup with the goal of improving the network's performance on the main task(s). We will introduce auxiliary tasks for sequence tagging constituent parsing later on in this work. A \textsc{mtl} architecture can also rely on \emph{partial sharing} when the different tasks do not fully share the internal representations \cite{duong2015low,rei2017semi,ruder2017learning} and  recent work has also shown that hierarchical sharing (e.g. low-level task outputs used as input for higher-level ones) could be beneficial \cite{sogaard2016deep, sanh2018hierarchical}.

\subsection{Policy Gradient Fine-tuning}\label{section-policy-gradient}

Policy gradient (\textsc{pg}) methods are a class of reinforcement learning algorithms that directly learn a parametrized policy, by which an agent selects actions based on the gradient of a scalar performance measure with respect to the policy. Compared to other reinforcement learning methods, \textsc{pg} is well-suited to \textsc{nlp} problems due to its appealing convergence properties and effectiveness in high-dimensional spaces \cite{sutton2018reinforcement}. 

Previous work on constituent parsing has employed \textsc{pg} methods to mitigate the effect of exposure bias, finding that they function as a model-agnostic substitute for dynamic oracles \citep{fried2018policy}. Similarly, \newcite{le2017tackling} apply \textsc{pg} methods to \newcite{chen2014fast}'s transition-based dependency parser to reduce error propagation. In this work, we also employ \textsc{pg} to fine-tune models trained using supervised learning. However, our setting (sequence tagging) has a considerably larger action space than a transition parser. To deal with that, we will adopt a number of variance reduction and regularization techniques to make reinforcement learning stable.

\section{Methods}\label{section-methods}

We describe the methods introduced in this work, motivated by current limitations of existing sequence tagging models, which are first reviewed. The source code can be found as a part of \url{https://github.com/aghie/tree2labels}.
%The source code will be released shortly
%and a link will be provided with a forthcoming version of this manuscript.

\subsection{Motivation and Analysis} 
For brevity, we limit this analysis to the English Penn Treebank (\textsc{ptb}) \cite{marcus1993building}. We reproduced the best setup by \newcite{GomVilEMNLP2018},
which we are using as baseline, and run the model on the development set. 
We below show insights for the elements of the output tuple $(n_t,c_t,u_t)$, where
$n_t$ is the number of levels in common between $w_t$ and $w_{t+1}$, $c_t$ is the non-terminal symbol shared at that level, and $u_t$ is a leaf unary chain located at $w_t$.

\begin{figure}[hbtp]
\centering
\includegraphics[width=1\columnwidth]{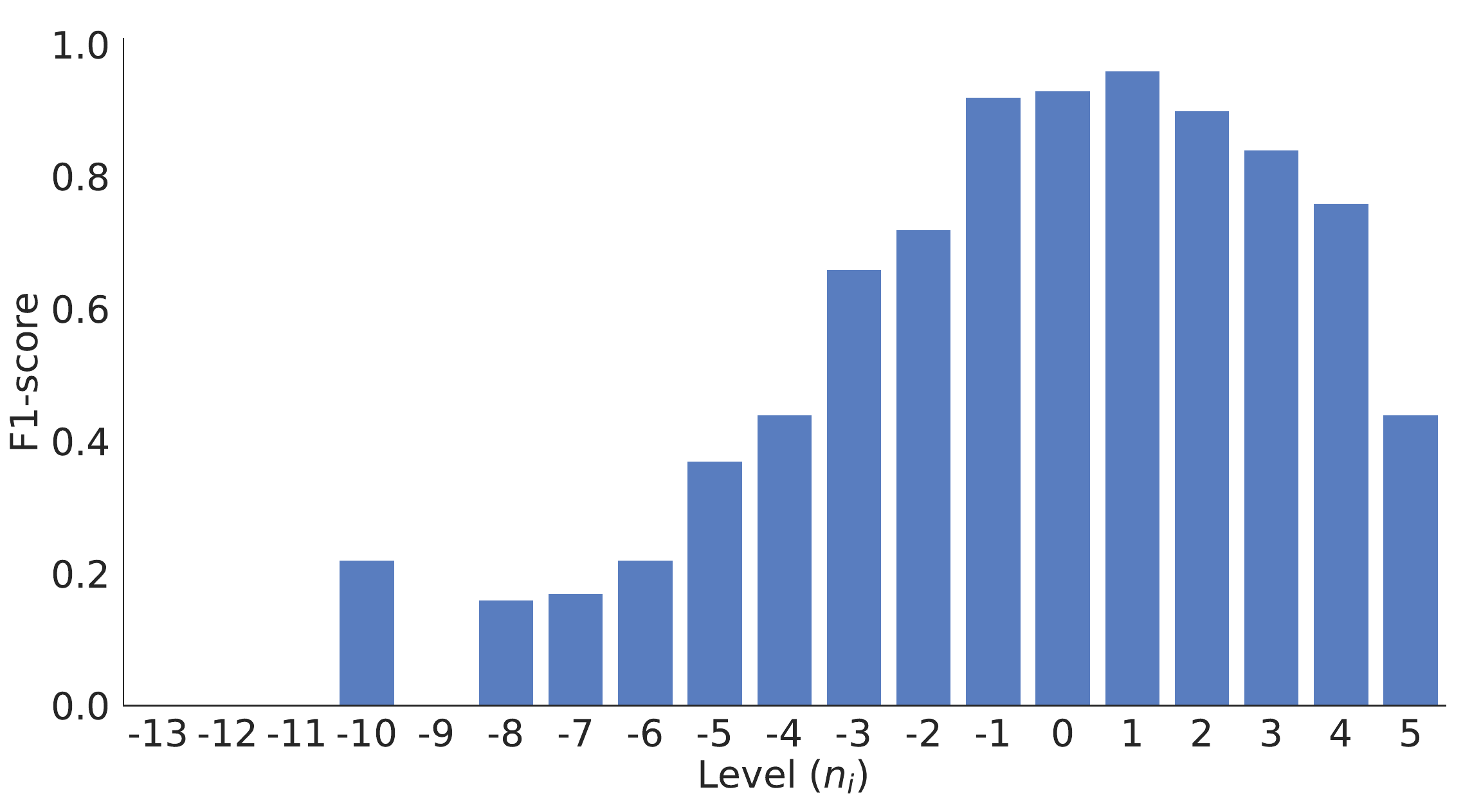}
\caption{\label{f-baseline-level-performance} F-score for $n_t$ labels on the \textsc{ptb} dev set using \newcite{GomVilEMNLP2018}.}
\end{figure}

\paragraph{High error rate on closing brackets} We first focus on predicting relative tree levels ($n_t$). See Figure \ref{f-baseline-level-performance} for F-scores over $n_t$ labels. The sparsity on negative $n_t$s is larger than for the positive ones, and we see that consequently, the performance is also significantly worse for negative $n_t$ values, and performance worsens with higher negative values. This indicates that the current model cannot effectively identify the end of long constituents. This is a known source of error for shift-reduce or chart-based parsers, but in the case of sequence tagging parsers, the problem seems particularly serious.

%Computed with 1423 computed with cut -f 3 ptb-train.seq_lu | sort -u | wc -l
% To measure the number labels that occur less than X times (e.g. 1) cut -f 3 ptb-train.seq_lu | sort | uniq -c | grep " 1 " | wc -l
\paragraph{Sparsity} The label space is large and sparse: the output labels are simply the possible values in the tuple $(n_t,c_t,u_t)$. An analysis over the \textsc{ptb} training set shows a total of 1423 labels, with 58\% of them occurring 5 or less times. These infrequent cases might be difficult to predict, even if some of the elements of the tuple are common. 

\paragraph{Greedy decoding} Greedy decoding is prone to issues such as error propagation. This is a known source of error in transition-based dependency parsing \cite{qi2017arc}; in contrast with graph-based parsing, in which parsing is reduced to global optimization  over edge-factored scores \cite{mcdonald2005non}. 

In the case of \textsc{bilstm}-based sequence tagging parsers, for a given word $w_t$, the output label as encoded by \newcite{GomVilEMNLP2018} only reflects a relation between $w_t$ and $w_{t+1}$. We hypothesize that even if the hidden vector representations are globally contextualized over the whole sequence, the intrinsic locality of the output label also turns into error propagation and consequently causes a drop in the performance. These hypotheses will be tested in \S \ref{section-experiments}. In particular, we will evaluate the impact of the different methods intended to perform structured inference (\S \ref{section-local-predictions}).

\subsection{Dynamic Encodings}

\newcite{GomVilEMNLP2018} encode the number of common ancestors $n_t$, from the output tuple $(n_t,c_t,u_t)$, as the variation with respect to $n_{t-1}$. We propose instead to encode certain elements of a sentence using a secondary linearization function. The aim is to generate a model that can dynamically switch between different tagging schemes at each time step $t$ to select the one that represents the relation between $w_t$ and $w_{t+1}$ in the most effective way. 

\begin{figure}[hbtp]
\centering
\includegraphics[width=1\columnwidth]{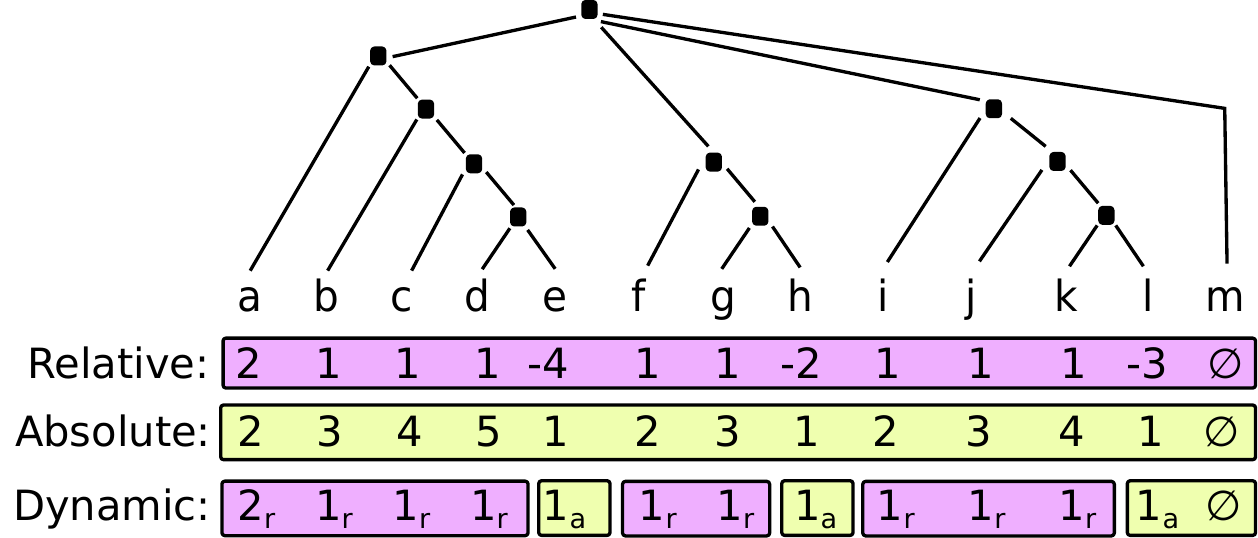}
\caption{\label{f-dynamic-example} A synthetic constituent tree where $n_t$ is encoded using a relative scheme, a top-down absolute scale, and an ideal dynamic combination. The relative scheme is  appropriate to open and close short constituents, but becomes sparse when encoding the large ones, e.g. $n_t$ for the tokens `e', `h' and 'l'. The opposite problem is observed for the top-down absolute scheme (e.g. tokens from `a' to `d').  The dynamic linearization combines the best of both encodings (we use the subscript `r' to denote the labels coming from the relative encoding, and `a' from the absolute one).}
\end{figure}

On the one hand, the relative-scale encoding is effective to predict the beginning and the end of short constituents, i.e. when a short constituent must be predicted ($|n_t| \leq 2$). On the other hand, with a relative encoding scheme, the F-score was low for words where the corresponding $n_t$ has a large negative value (as showed in Figure \ref{f-baseline-level-performance}). This matches a case where a long constituent must be closed: $w_t$ is located at a deep level in the tree and will only (probably) share a few ancestors with $w_{t+1}$. These configurations are encoded in a more sparse way by a relative scheme, as the $n_t$ value shows a large variability and it depends on the depth of the tree in the current time step. We can obtain a compressed representation of these cases by using a \emph{top-down absolute scale} instead, as any pair of words that share the same $m$ top levels will be equally encoded.
The absolute scale becomes however sparse when predicting deep levels.
Figure \ref{f-dynamic-example} illustrates the strengths and weaknesses of both encodings with an example, and how a dynamically encoded tree helps reduce variability on $n_t$ values.

In our particular implementation, we will be using the following setup:
\begin{itemize}
    \item $\Phi_{|w|}: T_{|w|} \rightarrow L^{|w|-1}$, the relative-scale encoding function, is used by default.
    
    \item $\Omega_{|w|}: T_{|w|} \rightarrow L'^{|w|-1}$ is the secondary linearization function that maps words to labels according to a top-down absolute scale. $\Omega$ is used iff: (1) $\Omega(w_{[t:t+1]})$ = $(n'_t,c'_t,u'_t)$ with $n'_t \leq 3$, i.e. $w_t$ and $w_{t+1}$ share at most the three top levels, and (2) $\Phi(w_{[t:t+1]})$ = $(n_t,c_t,u_t)$ with $n_t \leq -2 $, i.e. $w_{t}$ is at least located two levels deeper in the tree than $w_{t+1}$.\footnote{The values were selected based on the preliminary experiments of Figure \ref{f-baseline-level-performance}.}
\end{itemize}

\subsection{Decomposition of the label space}

We showed that labels of the form  $(n_t,c_t,u_t) \in L$ are sparse. An intuitive approach is to decompose the label space into three smaller sub-spaces, such that $n_i \in N$, $c_i \in C$ and $u_i \in U$. This reduces the output space from potentially $|N|\times|C|\times|U|$ labels to just $|N|+|C|+|U|$. We propose to learn this decomposed label space through a multi-task learning setup, where each of the subspaces is considered a different task, namely task$_N$, task$_C$ and task$_U$. The final loss is now computed as $\mathcal{L} = \mathcal{L}_n + \mathcal{L}_c + \mathcal{L}_u$.

We relied on a hard-sharing architecture, as it has been proved to reduce the risk of overfitting the shared parameters \cite{baxter1997bayesian}. A natural issue that arises is that the prediction of labels from different label sub-spaces could be interdependent to a certain extent, and therefore a hierarchical sharing architecture could also be appropriate. To test this, in preliminary experiments we considered variants of hierarchical sharing architectures. We fed the output of the task$_U$ as input to task$_N$ and/or task$_C$. Similarly, we tested whether it was beneficial to feed the output of task$_N$ into task$_C$, and viceversa. However, all these results did not improve those of the hard-sharing model. In this context, in addition to a generalization mechanism, the shared representation could be also acting as way to keep the model aware of the potential interdependencies that might exist between subtasks.

\subsection{Mitigating Effects of Greedy Decoding}\label{section-local-predictions}

We propose two ways to mitigate error propagation arising from greedy decoding in constituent parsing as sequence tagging: auxiliary tasks and policy gradient fine-tuning. Note that we want to optimize bracketing F-score {\em and} speed. For this reason we do {\em not}~explore approaches that come at a speed cost in testing time, such as beam-search or using conditional random fields \cite{Lafferty:ea:01} on top of our \textsc{lstm}. 

\paragraph{Auxiliary tasks} Auxiliary tasks force the model to take into account patterns in the input space that can be useful to solve the main task(s), but that remain ignored due to a number of factors, such as the distribution of the output label space \cite{rei2017semi}. In a similar fashion, we use auxiliary tasks as a way to force the parser to pay attention to aspects beyond those needed for greedy decoding. We propose and evaluate two separate strategies:
\begin{enumerate}
    \item Predict partial labels $n_{t+k}$ 
    that are $k$ steps from the current time step $t$. This way we can jointly optimize at each time step a prediction for the pairs $(w_t, w_{t+1})$, \dots, $(w_{t+k}, w_{t+k+1})$. In particular, we will experiment both with previous and upcoming $n_k$'s, setting $|k|$=$1$.\label{enu-aux-task-1}
    
    \item Predict the syntactic distances presented by \newcite{ShenDistance2018}, which reflect the order a sentence must be split to obtain its constituent tree using a top-down parsing algorithm \cite{stern2017minimal}.  The algorithm was initially defined for binary trees, but its adaptation to \emph{n}-ary trees is immediate: leaf nodes have a split priority of zero and the ancestors' priority is computed as the maximum priority of their children plus one. In this work, we use this algorithm in a sequence tagging setup: the label assigned to each token corresponds to the syntactic distance of the lowest common ancestor with the next token. This is illustrated in Figure \ref{f-syntactic-distances}. 
    
    \begin{figure}[hbtp]
    \centering
    \includegraphics[width=1\columnwidth]{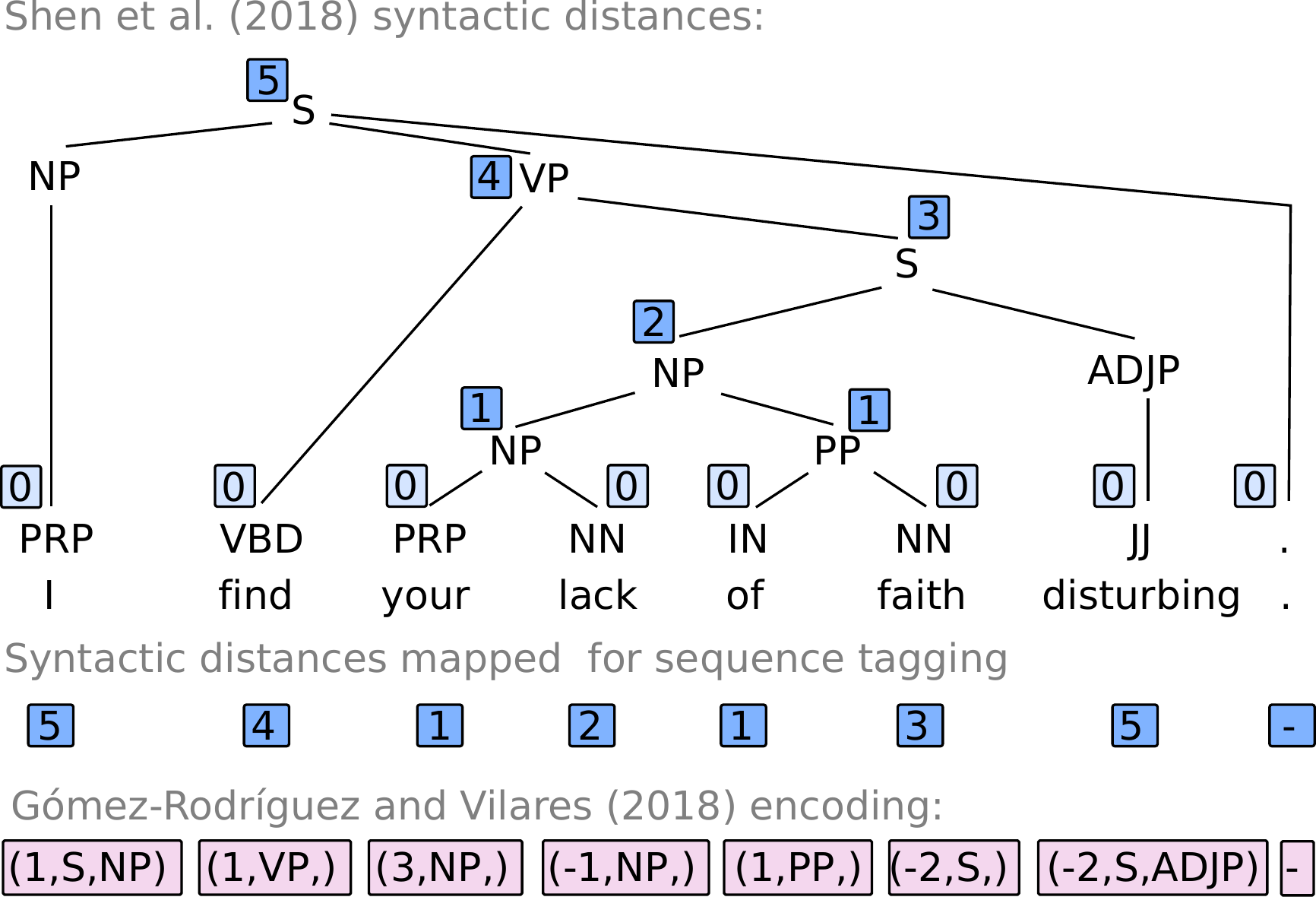}
    \caption{\label{f-syntactic-distances} A constituent with syntactic distances attached to each non-terminal symbol, according to \newcite{ShenDistance2018}. Distances can be used for sequence tagging, providing additional information to our base encoding \cite{GomVilEMNLP2018}}
    \end{figure}
    
\end{enumerate}

The proposed auxiliary tasks provide different types of contextual information. On the one hand, the encoding of the $n_t$s by \newcite{GomVilEMNLP2018} only needs to know about $w_t$ and $w_{t+1}$ paths to generate the label for the time step $t$. On the other hand, to compute the syntactic distance of a given non-terminal symbol, we need to compute the syntactic distances of its subtree, providing a more global, but also sparser context. For training, the loss coming from the auxiliary task(s) is weighted by $\beta$=0.1, i.e, the final loss is computed as $\mathcal{L} = \mathcal{L}_n + \mathcal{L}_c + \mathcal{L}_u + \beta\sum_{a}{\mathcal{L}_{a}} $.

\paragraph{Policy gradient fine-tuning}
Policy gradient training methods allow us to fine-tune our models with a tree-level objective, optimizing directly for bracketing F-score. We start off with a converged supervised model as our initial policy. The sequence labeling model can be seen as a functional approximation of the policy $\pi$ parametrized by $\theta$, which at timestep $t$ selects a label $l_t$=$(n_t,c_t,u_t)$\footnote{3 different labels in the \textsc{mtl} setting.} given the current state of the model's parameters, $s_t$. 
The agent's reward, $R_{tree}$, is then derived from the bracketing F-score. 
This can be seen as a variant of the \textsc{reinforce} algorithm \citep{williams1992simple} where the policy is updated by gradient ascent in the direction of:

\begin{equation}
    \Delta_\theta log\pi(l_t|s_t; \theta)R_{tree}
\end{equation}

\paragraph{Baseline and Variance Reduction}
We use as baseline a copy of a pre-trained model where the parameters are frozen.
The reward used to scale the policy gradient can then be seen as an estimate of the advantage of an action $l_t$ in state $s_t$ over the baseline model. This is equivalent to $R_{tree} - B_{tree}$, where $R_{tree}$ is the bracketing F-score of a sequence sampled from the current policy and $B_{tree}$ is the the tree-level F-score of the sequence greedily predicted by the baseline. To further reduce the variance, we standardize the gradient estimate $\Delta_\theta$ using its running mean and standard deviation for all candidates seen in training so far. In initial experiments without these augmentations, we observed that fine-tuning with vanilla \textsc{pg} 
often led to a deterioration in performance. To encourage exploration away from the converged supervised model's policy, we add the entropy of the policy to the objective function  \citep{williams1991function}. Moreover, following \newcite{lillicrap2015continuous}, we optionally add noise sampled from a noise process $N$ to the policy. The gradient
of our full fine-tuning objective function takes the following form:

\begin{multline}
    \Delta_\theta (log\pi(l_t|s_t; \theta) + N) (R_{tree} - B_{tree}) \\ + \beta  \Delta_\theta H(\pi(s_t; \theta) + N)
\end{multline}

\noindent where $H$ is the entropy and $\beta$ controls the strength of the entropy regularization term.

\section{Experiments}\label{section-experiments}

We now review the impact of the proposed techniques on a wide variety of settings.

\paragraph{Datasets} We use the English Penn Treebank (\textsc{ptb}) \cite{marcus1993building} and the Chinese Penn Treebank (\textsc{ctb}) \cite{xue2005penn}. For these, we use the same predicted PoS tags as  \newcite{DyerRecurrent2016}. We also provide detailed results on the \textsc{spmrl} treebanks \cite{seddah2014introducing},\footnote{Except for Arabic, for which we do not have the license.}  a set of datasets for constituent parsing on morphologically rich languages. For these, we use the predicted PoS tags provided together with the corpora. To the best of our knowledge, we provide the first evaluation on the \textsc{spmrl} datasets for sequence tagging constituent parsers.

\paragraph{Metrics} We report bracketing F-scores, using the \textsc{evalb} and the \textsc{eval-spmrl} scripts parametrized with the \textsc{collins}.prm and spmrl.prm files, respectively. We measure the speed in terms of sentences per second.

\paragraph{Setup} We use \textsc{ncrf}pp \cite{yang2017ncrf}, for direct comparison against \citet{GomVilEMNLP2018}. We adopt bracketing F-score instead of label accuracy for model selection and report this performance as our second baseline. After 100 epochs, we select the model that fared best on the development set.
We use GloVe embeddings \cite{pennington2014glove} for our English models and \texttt{zzgiga} embeddings \cite{Liu2017InOrder}  for the Chinese models, for a more homogeneous comparison against other parsers \cite{DyerRecurrent2016,Liu2017InOrder,Fer2018Faster}. ELMo \cite{peters2018deep} or BERT \cite{devlin2018bert} could be used to improve the precision, but in this paper we focus on keeping a good speed-accuracy tradeoff. For \textsc{spmrl}, no pretrained embeddings are used, following \newcite{KitaevConstituencyACL2018}. As a side note, if we wanted to improve the performance on these languages we could rely on the CoNLL 2018 shared task pretrained word embeddings \cite{zeman2018conll} or even the multilingual \textsc{bert} model\footnote{\url{https://github.com/google-research/bert/blob/master/multilingual.md}}. Our models are run on a single CPU\footnote{Intel Core i7-7700 CPU 4.2 GHz} (and optionally on a consumer-grade GPU for further comparison) using a batch size of 128 for testing. Additional hyperparameters can be found in Appendix \ref{appendix-a}.

\subsection{Results}

Table \ref{table-ptb-dev-set} contrasts the performance of our models against the baseline on the \textsc{ptb} development set.

\begin{table}[htbp]
\tabcolsep=0.2cm
\begin{center}
\small
\begin{tabular}{lccc}
 \hline
 \bf Model & \bf F-score & \bf (+/-)  & \bf Sents/s \\
 \hline
 \defcitealias{GomVilEMNLP2018}{G\'omez and Vilares (2018)}\citetalias{GomVilEMNLP2018} &90.60&-&109\\
 Our baseline &90.64&(+0.04)&111\\
 + \textsc{de} & 91.16 &(+0.56)& 111  \\
 + \textsc{mtl} & 91.27 &(+0.67)& 130 \\
 \hline
 aux($n_{t+1}$)& 90.19 &(+0.59)&130\\
 aux($n_{t-1}$) & 91.40&(+0.80)&130\\
 aux(distances)  & 91.44 &(+0.84)&130\\
  \hline
 + \textsc{pg}&\bf{91.67}&(+1.07)&130\\
 \hline

\end{tabular}
\end{center}
\caption{\label{table-ptb-dev-set} Results on the \textsc{ptb} dev set, compared against \newcite{GomVilEMNLP2018}. \textsc{de} refers to dynamic encoding and \textsc{mtl} to a model that additionally casts the problem as multi-task learning. Each auxiliary task is added separately to the baseline with \textsc{de} and \textsc{mtl}. Policy gradient fine-tunes the model that includes the best auxiliary task.}
\end{table}

%BEFORE CORRIGENDUM
% \begin{table}[htbp]
% \tabcolsep=0.2cm
% \begin{center}
% \small
% \begin{tabular}{lccc}
%  \hline
%  \bf Model & \bf F-score & \bf (+/-)  & \bf Sents/s \\
%  \hline
%  \defcitealias{GomVilEMNLP2018}{G\'omez and Vilares (2018)}\citetalias{GomVilEMNLP2018} &89.70&-&109\\
%  Our baseline &89.77&(+0.07)&111\\
%  + \textsc{de} & 90.22 &(+0.52)& 111  \\
%  + \textsc{mtl} & 90.38 &(+0.68)& 130 \\
%  \hline
%  aux($n_{t+1}$)& 90.41 &(+0.71)&130\\
%  aux($n_{t-1}$) & 90.57&(+0.87)&130\\
%  aux(distances)  & 90.55 &(+0.85)&130\\
%   \hline
%  + \textsc{pg}&\bf{90.70}&(+1.00)&130\\
%  \hline

% \end{tabular}
% \end{center}
% \caption{\label{table-ptb-dev-set} Results on the \textsc{ptb} dev set, compared against \newcite{GomVilEMNLP2018}. \textsc{de} refers to dynamic encoding and \textsc{mtl} to a model that additionally casts the problem as multi-task learning. Each auxiliary task is added separately to the baseline with \textsc{de} and \textsc{mtl}. Policy gradient fine-tunes the model that includes the best auxiliary task.}
% \end{table}

\begin{table*}[!]
\begin{center}
\small
\begin{tabular}{l|c|cccccccc}
\hline
 \bf Model & \bf \textsc{ctb} & \bf Basque & \bf French & \bf German & \bf Hebrew & \bf Hungarian & \bf Korean & \bf Polish & \bf Swedish  \\
 \hline
Our baseline& 88.57&87.62&80.19&86.48&89.09&88.61&82.79&92.60&78.82\\
\textsc{+de}&88.37&87.57&80.27&87.44&88.82&88.42&83.11&93.35&78.24\\
\textsc{+mtl}& 88.57&89.12&80.84&87.54&\bf 92.63& 89.55&  83.27& 93.81& 81.71\\
\hline
aux($n_{t+1}$)&88.73& 89.37&81.09&87.59&92.57&89.50& 83.28&93.86&81.70\\
aux($n_{t-1}$)&88.48& 89.19&80.91&87.67&92.45&89.52&83.37& 93.87&81.61\\
aux(distances)&88.51& 89.23& 81.17& 87.68& 92.56& 89.58& 83.39&93.83& 82.02\\
\hline
\textsc{+pg}&\bf 89.01&\bf 89.44&\bf 81.28&\bf87.83&92.56&\bf89.63&\bf83.63&\bf93.93&\bf82.05\\
\hline

\end{tabular}
\end{center}
\caption{\label{table-other-dev-sets} Results on the \textsc{ctb} and \textsc{spmrl} dev sets}
\end{table*}

%BEFORE CORRIGENDUM
% \begin{table*}[!]
% %\tabcolsep=0.15cm
% \begin{center}
% \small
% \begin{tabular}{l|c|cccccccc}
% \hline
%  \bf Model & \bf \textsc{ctb} & \bf Basque & \bf French & \bf German & \bf Hebrew & \bf Hungarian & \bf Korean & \bf Polish & \bf Swedish  \\
%  \hline
% Our baseline& 88.57&87.93&81.09&87.83&89.27&88.85&83.51&92.60&80.11\\
% \textsc{+de}&88.37&87.91&81.16&\bf 88.81&89.03&88.70&83.92&93.35&79.57\\
% \textsc{+mtl}& 88.57&89.41&81.70&88.52&\bf 92.72& 89.73&  84.10& 93.81& 82.83\\
% \hline
% aux($n_{t+1}$)&88.73& 89.65&81.95&88.64&92.65&89.69& 84.09&93.86&82.82\\
% aux($n_{t-1}$)&88.48& 89.47&81.77&88.58&92.53&89.71&84.13& 93.87&82.74\\
% aux(distances)&88.51& 89.48& 82.02& 88.68& 92.66& 89.80& 84.20&93.83& 83.12\\
% \hline
% \textsc{+pg}&\bf 89.01&\bf 89.73&\bf 82.13&88.80&92.66&\bf89.86&\bf84.45&\bf93.93&\bf83.15\\
% \hline

% \end{tabular}
% \end{center}
% \caption{\label{table-other-dev-sets} Results on the \textsc{ctb} and \textsc{spmrl} dev sets}
% \end{table*}

\begin{figure}[hbtp]
\centering
\includegraphics[width=1\columnwidth]{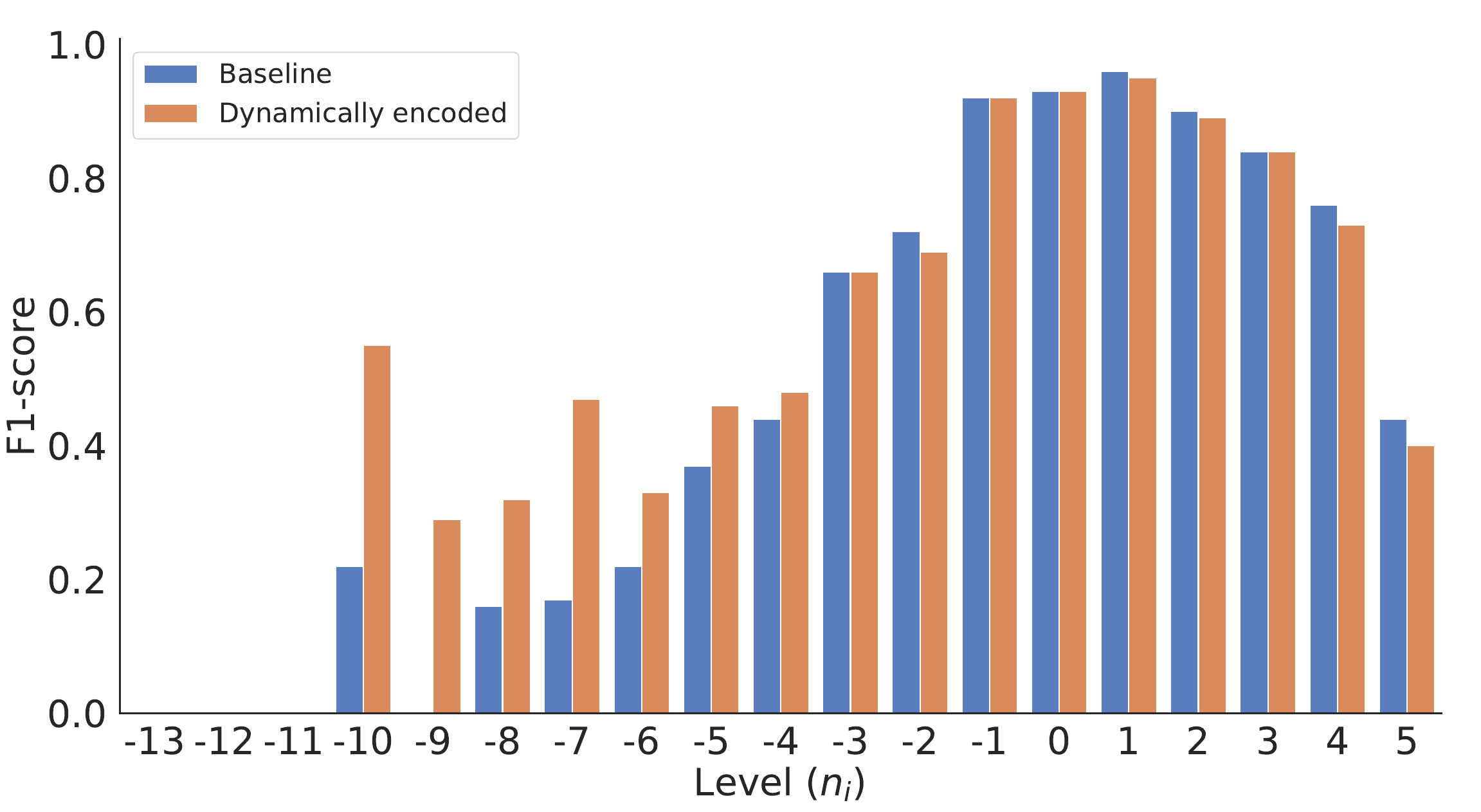}
\caption{\label{f-baseline-level-performance-2} F-score for $n_t$s on the \textsc{ptb} dev set, obtained by the \newcite{GomVilEMNLP2018} baseline (in blue, first bar for each $n_t$, already shown in Figure \ref{f-baseline-level-performance}) and our model with dynamically encoded trees (in orange, second bar).}
\end{figure}

To show that the model which employs dynamic encoding is better (+0.56) than the baseline when it comes to closing brackets from long constituents, we compare their F-scores in Figure \ref{f-baseline-level-performance-2}.
When we recast the constituent-parsing-as-sequence-tagging problem as multi-task learning, we obtain both a higher bracketing F-score (+0.67) and speed (1.17x faster). 
Fusing strategies to mitigate issues from greedy decoding also leads to better models (up to +0.84 when adding an auxiliary task\footnote{We observed that adding more than one auxiliary task did not translate into a clear improvement. We therefore chose the auxiliary task that performed the best in the development set.} and up to +1.07 if we also fine-tune with \textsc{pg}). Note that including auxiliary tasks and \textsc{pg} come at a time cost in training, but not in testing, which makes them suitable for fast parsing.

%BEFORE CORRIGENDUM
%To show that the model which employs dynamic encoding is better (+0.52) than the baseline when it comes to closing brackets from long constituents, we compare their F-scores in Figure \ref{f-baseline-level-performance-2}.When we recast the constituent-parsing-as-sequence-tagging problem as multi-task learning, we obtain both a higher bracketing F-score (+0.68) and speed (1.17x faster). Fusing strategies to mitigate issues from greedy decoding also leads to better models (up to +0.87 when adding an auxiliary task\footnote{We observed that adding more than one auxiliary task did not translate into a clear improvement. We therefore chose the auxiliary task that performed the best in the development set.} and up to +1.00 if we also fine-tune with \textsc{pg}). Note that including auxiliary tasks and \textsc{pg} come at a time cost in training, but not in testing, which makes them suitable for fast parsing.

Table \ref{table-other-dev-sets} replicates the experiments on the \textsc{ctb} and the \textsc{spmrl} dev sets. The dynamic encoding improves the performance of the baseline on large treebanks, e.g. German, French or Korean, but causes some drops in the smaller ones, e.g. Swedish or Hebrew. Overall, casting the problem as multitask learning and the strategies used to mitigate error propagation lead to improvements.

For the experiments on the test sets we select the models that summarize our contributions: the models with dynamic encoding and the multi-task setup, the models including the best auxiliary task, and the models fine-tuned with policy gradient.

\begin{table}[hbtp]
\tabcolsep=0.07cm
\begin{center}
\small
\begin{tabular}{lrrc}
\hline
 \bf Model & \bf Sents/s&\bf Hardware & \bf F-score\\
 \hline
 \newcite{vinyals2015grammar}&120&Many CPU&{88.30}\\
 \newcite{CoaCra2016}&168&1 CPU&88.60\\
 \defcitealias{Fer2015Parsing}{Fern\'andez and Martins (2018)}\citetalias{Fer2015Parsing}&41&1 CPU&90.20\\
 \newcite{zhu2013fast}&90&1 CPU&90.40\\
 \newcite{DyerRecurrent2016}&17&1 CPU&91.20\\
 \newcite{stern2017minimal}&76&16 CPU&91.77\\
 \newcite{ShenDistance2018}&111&1 GPU&91.80\\
  \newcite{KitaevConstituencyACL2018}&213&2 GPU&93.55\\
 (single model)&&&\\
 \newcite{KitaevConstituencyACL2018}&71&2 GPU&95.13\\
 (with \textsc{elm}o)&&&\\
 \newcite{Kitaev2018BERT}&-&-&\emph{95.77}\\
 (ensemble and \textsc{bert})&&&\\
  \defcitealias{GomVilEMNLP2018}{G\'omez and Vilares (2018)}\citetalias{GomVilEMNLP2018}&115& 1 CPU&90.70\\
  \hline
 Our baseline &115&1 CPU& 90.75\\
 \textsc{+de} &115&1 CPU& 90.85  \\
 \textsc{+mtl} &132&1 CPU&90.97\\
 + best aux &132&1 CPU&90.97\\
\textsc{+pg}&132&1 CPU&91.13\\
 \textsc{+pg}&942&1 GPU&91.13\\
 \shadowed{\textsc{+pg} (no char emb)}&\shadowed{149}&\shadowed{1 CPU}&\shadowed{91.09}\\
 \shadowed{\textsc{+pg} (no char emb)}&\shadowed{1267}&\shadowed{1 GPU}&\shadowed{91.09}\\
\hline

\end{tabular}
\end{center}
\caption{\label{table-sota-ptb} Comparison on the \textsc{ptb} test set. \newcite{Kitaev2018BERT} are results published after this work was submitted (italics represent the cases where they obtain a new state of the art on the corresponding language).}
\end{table}

\begin{table}[bpth]
\begin{center}
\small
\begin{tabular}{lc}
\hline
 \bf Model & \bf F-score   \\
 \hline
 \newcite{zhu2013fast}&83.2\\
 \newcite{DyerRecurrent2016}&84.6\\
 \newcite{Liu2017InOrder}&86.1\\
 \newcite{ShenDistance2018}&86.5\\
 \defcitealias{Fer2018Faster}{Fern\'andez and G\'omez-Rodr\'iguez (2018)}\citetalias{Fer2018Faster}&86.8\\
 \defcitealias{GomVilEMNLP2018}{G\'omez and Vilares (2018)}\citetalias{GomVilEMNLP2018}&84.1\\
\hline
 Our baseline&83.90\\
 +\textsc{de}&83.98\\
 +\textsc{mtl}&84.24\\
 +best aux&85.01\\
 \textsc{+pg}&85.61\\
 \shadowed{\textsc{+pg} (no char emb)}&\shadowed{83.93}\\
 \hline
\end{tabular}
\end{center}
\caption{\label{table-sota-chinese} Comparison on the \textsc{ctb} test set}
\end{table}

\begin{table*}[!]
\tabcolsep=0.07cm
\begin{center}
\small
\begin{tabular}{p{5.3cm}|ccccccccc}
 \hline
 \bf Model & \bf Basque & \bf French & \bf German & \bf Hebrew & \bf Hungarian & \bf Korean & \bf Polish & \bf Swedish &\bf Avg \\
  \hline
\newcite{Fer2015Parsing}&85.90&78.75&78.66&88.97&88.16&79.28&91.20&82.80&84.21\\
 \newcite{CoaCra2016}&86.24&79.91&80.15&88.69&90.51&85.10&92.96&81.74&85.67\\
 \newcite{bjorkelund2014introducing} (ensemble)&88.24&82.53&81.66&89.80&91.72&83.81&90.50&85.50&86.72\\
 \newcite{coavoux2017multilingual}&88.81&82.49&85.34&89.87&92.34&86.04&93.64&84.00&87.82 \\
 \newcite{KitaevConstituencyACL2018}&89.71&\bf 84.06&\bf 87.69&90.35&\bf 92.69&\bf 86.59&93.69&84.35&\bf88.64 \\
 \newcite{Kitaev2018BERT} (with \textsc{bert})&\emph{91.63}&\emph{87.42}&\emph{90.20}&\emph{92.99}&\emph{ 94.90}&\emph{88.80}&\emph{96.36}&\emph{88.86}&\emph{91.40}\\
 \hline
Baseline&89.20&79.58&82.33&88.67&90.10&82.63&92.48&82.40&85.92\\
\textsc{+de}&89.19&79.72&82.91&88.60&89.65&82.86&93.20&82.11&86.03\\
\textsc{+mtl}&90.60&80.02&83.48&91.91&90.32&83.11&93.80&85.19&87.30\\
+best aux&\bf 90.91&80.33&83.49&\bf92.05&90.33&82.97&93.84&\bf 85.58&87.44\\
\textsc{+pg}&90.85&80.40&83.42&\bf92.05&90.38&83.24&\bf93.93&85.54&87.48\\
\shadowed{\textsc{+pg} (no char emb)}&\shadowed{89.81}&\shadowed{80.41}&\shadowed{83.60}&\shadowed{91.75}&\shadowed{90.01}&\shadowed{82.65}&\shadowed{93.87}&\shadowed{85.46}&\shadowed{87.20}\\
\hline

\end{tabular}
\end{center}
\caption{\label{table-sota-spmrl} Comparison on the test \textsc{spmrl} datasets (except Arabic). \newcite{Kitaev2018BERT} are results published after this work was submitted (italics represent the cases where they obtain a new state of the art on the corresponding language).}
\end{table*}

Tables \ref{table-sota-ptb}, \ref{table-sota-chinese} and \ref{table-sota-spmrl} compare our parsers against the state of the art on the \textsc{ptb}, \textsc{ctb} and \textsc{spmrl} test sets.
\newcite{GomVilEMNLP2018} also run experiments without character embeddings, to improve speed without suffering from a big drop in performance. For further comparison, we also include them as additional results (shadowed). In a related line, \newcite{smith2018investigation} show that for dependency parsing two out of three embeddings (word, postag and characters) can suffice.
\subsection{Discussion}

The results across the board show that the dynamic encoding has a positive effect on 7 out of 10 treebanks. Casting the constituent-parsing-as-sequence-labeling problem as \textsc{mtl} surpasses the baseline for all tested treebanks (and it leads to better parsing speeds too). Finally, by mitigating issues from greedy decoding we further improve the performance of all models that include dynamic encodings and multi-task learning.

On the \textsc{ptb}, our models are both faster and more accurate than existing sequence tagging or sequence-to-sequence models, which already were among the fastest parsers \cite{GomVilEMNLP2018,vinyals2015grammar}. We also outperform other approaches that were not surpassed by the original sequence tagging models in terms of F-score \cite{zhu2013fast,Fer2015Parsing}. On the \textsc{ctb} our techniques also have a positive effect. The baseline parses 70 sents/s on the \textsc{ctb}, while the full model processes up to 120. The speed up is expected to be larger than the one obtained for the \textsc{ptb} because the size of the label set for the baseline is bigger, and it is reduced in a greater proportion when the constituent-parsing-as-sequence-labeling problem is cast as \textsc{mtl}.

On the \textsc{spmrl} corpora, we provide the first evaluation of sequence labeling constituent parsers, to verify if these perform well on morphologically rich languages. We then evaluated whether the proposed techniques can generalize on heterogeneous settings.
The tendency observed for the original tagging models by \newcite{GomVilEMNLP2018} is similar to the one for the \textsc{ptb} and \textsc{ctb}: they improve other fast parsers, e.g. \newcite{CoaCra2016}, in 3 out of 8 treebanks and \newcite{Fer2015Parsing} in 6 out of 8, but their performance is below more powerful models. When incorporating the techniques presented in this work, we outperform the original sequence tagging models on all datasets. We outperform the current best model for Basque, Hebrew and Polish \cite{KitaevConstituencyACL2018} and for Swedish \cite{bjorkelund2014introducing}, which corresponds to the four smallest treebanks among the \textsc{spmrl} datasets. This indicates that even if sequence tagging models are conceptually simple and fast, they can be very suitable when little training data is available. This is also of special interest in terms of research for low-resource languages. Again, casting the problem as \textsc{mtl} reduces the parsing time for all tested treebanks, as reflected in Table \ref{table-spmrl-speeds}. Finally, for treebanks such as French, designing methods to handle multi-word expressions could lead to better results, getting closer to other parsers \cite{coavoux2017multilingual}.

\begin{table}[bpth]
\tabcolsep=0.16cm
\begin{center}
\small
\begin{tabular}{l|ccc}
\hline
 \multirow{2}{*}{\bf Dataset} &{\bf Baseline} &{\bf Full} &\bf  \shadowed{\bf Full (no char)} \\
 &{\bf speed} &\bf speed\textsubscript{(increase)}&\bf \shadowed{ speed\textsubscript{(increase)}}\\
 \hline
 Basque&179&223\textsubscript{ (1.25x)}&\shadowed{257}\textsubscript{ \shadowed{(1.44x)}}\\
 French&76&91\textsubscript{ (1.20x)}&\shadowed{104}\textsubscript{ \shadowed{(1.37x)}}\\
 German&70&100\textsubscript{ (1.43x)}&\shadowed{108}\textsubscript{ \shadowed{(1.54x)}}\\
 Hebrew&44&102\textsubscript{ (2.32x)}&\shadowed{115}\textsubscript{ \shadowed{(2.61x)}}\\
 Hungarian&93&134\textsubscript{ (1.44x)}&\shadowed{150}\textsubscript{ \shadowed{(1.61x)}}\\
 Korean&197&213\textsubscript{ (1.08x)}&\shadowed{230}\textsubscript{ \shadowed{(1.17x)}}\\
 Polish&187&253\textsubscript{ (1.35x)}&\shadowed{278}\textsubscript{ \shadowed{(1.49x)}}\\
 Swedish&98&158\textsubscript{ (1.61x)}&\shadowed{187}\textsubscript{ \shadowed{(1.81x)}}\\

\hline

\end{tabular}
\end{center}
\caption{\label{table-spmrl-speeds} Comparison of speeds on the \textsc{spmrl} datasets}
\end{table}

\section{Conclusion}

We have explored faster and more precise sequence tagging models for constituent parsing. We proposed a multitask-learning architecture that employs dynamic encodings, auxiliary tasks, and policy gradient fine-tuning. We performed experiments on the English and Chinese Penn Treebanks, and also on the \textsc{spmrl} datasets. Our models improve current sequence tagging parsers on all treebanks, both in terms of performance and speed. We also report state-of-the-art results for the Basque, Hebrew, Polish, and Swedish datasets. The methods presented in this work are specifically designed for constituent parsing. However, it seems natural to apply some of these to other \textsc{nlp} tagging tasks, e.g. using multi-task learning to predict sub-level morphological information for morphologically-rich part-of-speech tagging.

\section*{Acknowlegments}

DV has received support from the European Research Council (ERC), under the European Union's Horizon 2020 research and innovation programme (FASTPARSE, grant agreement No 714150), from the TELEPARES-UDC project
(FFI2014-51978-C2-2-R) and the ANSWER-ASAP project (TIN2017-85160-C2-1-R) from MINECO, and from Xunta de Galicia (ED431B 2017/01). MA and AS are funded by a  Google Focused Research Award.

\bibliographystyle{acl_natbib}
\bibliography{naaclhlt2019}

%\clearpage
\appendix

\section{Appendices}\label{appendix-a}

For the \textsc{bilstm}-based model, we essentially follow the configuration of the baseline \cite{GomVilEMNLP2018} for an homogenous comparison. We detail the hyperparameters in Table \ref{table-appendix-gomvil}.\footnote{Note that the noise sampling is only used for Swedish in the final models based on development set results with and without it.}

\makeatletter
\setlength{\@fptop}{0pt}
\makeatother

\begin{table}[!th]
\begin{center}
\small
\begin{tabular}{p{3.6cm}r}
\hline
 \bf Hyperparameter & \bf Value \\
 \hline
 \textsc{bilstm} size&800 \\
 \# \textsc{bilstm} layers& 2\\
 optimizer&\textsc{sgd}\\
 loss& cat. cross-entropy\\
 learning rate&0.2\\
 decay (linear)&0.05\\
 momentum&0.9\\
 dropout&0.5\\
 word emb size&100\\
 features size&20\\
 character emb size&30\\
 batch size training&8 \\
 training epochs&100\\
 batch size test&128 \\
 \hline
 \bf PG finetuning Hyperparameter& \bf Value \\
  \hline
 \# samples & 8 \\
 learning rate&0.0005 \\
 entropy regularization coefficient & 0.01 \\
 variance reduction burn-in \# of examples & 1000 \\
 layers frozen & word \& char embeddings \\ 
 noise process initial stddev & 0.1 \\
 noise process desired action stddev & 0.5 \\
 noise process adaptation coefficient & 1.05 \\
 \hline
 
\hline

\end{tabular}
\end{center}
\caption{\label{table-appendix-gomvil} Additional hyperparameters of the base model and Policy Gradient fine-tuning}
\end{table}

\twocolumn[{%
 \centering
 \Large\bf Corrigendum to Better, Faster, Stronger Sequence Tagging Constituent Parsers\\[1.5cm]}]

\begin{abstract}

Due to an implementation bug in the evaluation, the \textsc{evalb} scripts were not parametrized by the \textsc{collins} and spmrl parameter files. This corrigendum describes the changes that this has caused with respect to the original version, which still can be downloaded from: \url{https://arxiv.org/abs/1902.10985v2}. 

\end{abstract}

\section*{Results after correction}

\noindent \textbf{Note:} For model selection, we still do not exclude any non-terminal or pre-terminal from the evaluation, while for official comparison on the dev and test sets we now use the \textsc{collins}.prm and spmrl.prm files to parametrize the \textsc{evalb} scripts.\\

\noindent This corrected version contains improved results for the experiments on the \textsc{ptb}, as the \textsc{collins}.prm file excludes from the evaluation some pre-terminals related to punctuation. For the experiments in the \textsc{spmrl} datasets, punctuation is taken into account, but the non-terminals \textsc{top, s1, root, vroot} are stripped off when using the spmrl.prm parameter file. This translates into lower results ($\sim$0.6 points on average), but the tendencies showed in the paper still hold.\\

\noindent With respect to the experiments with the full models, we were relying on the models trained with the auxiliary task that performed the best on the development set. Although differences across auxiliary tasks were in general small; for most of the treebanks the auxiliary task that performed the best with the buggy evaluation still keeps to do so with the corrected one. There are two exceptions where the ranking of the top auxiliary task change by a tiny difference: English (0.02) and Hebrew (0.01). For these models, we re-trained and updated the full models accordingly. 

\end{document}